%% file: main.tex
\documentclass[runningheads]{llncs}
\usepackage{graphicx}
\usepackage{tikz}
\usepackage{comment}
\usepackage{amsmath,amssymb} % define this before the line numbering.
\usepackage{color}
\usepackage{url}
\usepackage{hyperref}
\usepackage{wrapfig}

\begin{document}
% \linenumbers
\pagestyle{headings}
\mainmatter
\def\ECCVSubNumber{761}  % Insert your submission number here

\title{Joint Learning of Social Groups, Individuals Action and Sub-group Activities in Videos} % Replace with your title

%\begin{comment}
%\titlerunning{ECCV-20 submission ID \ECCVSubNumber} 
%\authorrunning{ECCV-20 submission ID \ECCVSubNumber} 
%\author{Anonymous ECCV submission}
%\institute{Paper ID \ECCVSubNumber}
%\end{comment}
%******************

%\begin{comment}
\titlerunning{Joint Learning of Social Groups, Ind. Action and Sub-group Act. in Videos}
\author{Mahsa Ehsanpour\inst{1,3} \and
Alireza Abedin\inst{1} \and
Fatemeh Saleh\inst{2,3} \and Javen Shi\inst{1} \and \\ Ian Reid\inst{1,3} \and Hamid Rezatofighi\inst{1}}
\authorrunning{M. Ehsanpour et al.}
\institute{The University of Adelaide \and
Australian National University \and Australian Centre for Robotic Vision\\
\email{mahsa.ehsanpour@adelaide.edu.au}}
%\end{comment}
%******************
\maketitle
\begin{abstract}
The state-of-the art solutions for human activity understanding from a video stream formulate the task as a spatio-temporal problem which requires joint localization of all individuals in the scene and classification of their actions or group activity over time. 
Who is interacting with whom, e.g. not everyone in a queue is interacting with each other, is often not predicted. There are scenarios where people are best to be split into sub-groups, which we call social groups, and each social group may be engaged in a different social activity. In this paper, we solve the problem of simultaneously grouping people by their social interactions, predicting their individual actions and the social activity of each social group, which we call the social task. Our main contributions are: i) we propose an end-to-end trainable framework for the social task; ii) our proposed method also sets the state-of-the-art results on two widely adopted benchmarks for the traditional group activity recognition task~(assuming individuals of the scene form a single group and predicting a single group activity label for the scene); iii) we introduce new annotations on an existing group activity dataset, re-purposing it for the social task. The data and code for our method is publicly available \footnote{https://github.com/mahsaep/Social-human-activity-understanding-and-grouping}.
\end{abstract}

\input{intro}

\input{relatedwork}
\input{method}

\input{dataset}
\input{experiment}

\input{conclusion}
\bibliographystyle{splncs04}
\bibliography{main}
\end{document}

%% file: intro.tex
\section{Introduction}
%\vspace{-.5em}
Recognising individuals' action and group activity from video is a widely studied problem in computer vision. This is crucial for surveillance systems, autonomous driving cars and robot navigation in environments where humans are present~\cite{caba2015activitynet,collins2000introduction,kruse2013human}. In the last decade, most effort from the community has been dedicated to extract reliable spatio-temporal representations from video sequences. Dominantly, this was investigated in a simplified scenario where each video clip was trimmed to involve a single action, hence a classification problem~\cite{carreira2017quo,kuehne2011hmdb,simonyan2014two,soomro2012ucf101}. 
Recently, the task of human activity understanding has been extended to more challenging and realistic scenarios, where the video is untrimmed and may include multiple individuals performing different actions~\cite{girdhar2019video,gu2018ava,meva,sun2018actor}. In parallel, there are works focusing on predicting a group activity label to represent a collective behaviour of all the actors in the scene~\cite{choi2011learning,deng2015deep,lan2011discriminative}. There are also independent works aiming at only grouping individuals in the scene based on their interactions \cite{choi2014discovering,ge2012vision,hung2011detecting,setti2013multi,swofford2019dante} or joint inferring of groups, events and human roles in aerial videos \cite{shu2015joint} by utilizing hand-crafted features.
\begin{figure}[t]
  \centering
  \begin{minipage}[b]{.35\linewidth} 
		\includegraphics[clip,width=1.0\linewidth]{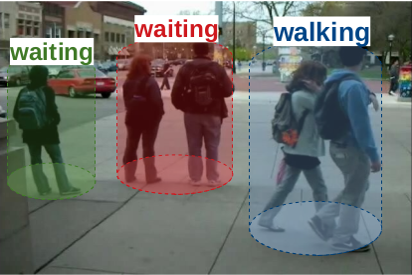}
		%\centerline{(a)}\medskip
	\end{minipage}
	\begin{minipage}[b]{.35\linewidth}
		\includegraphics[width=1.0\linewidth]{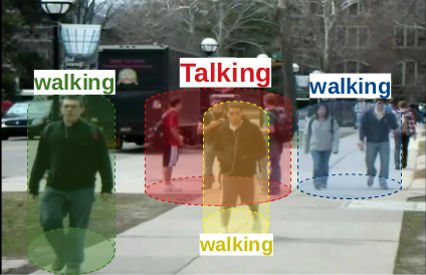}
		%\centerline{(b)}\medskip
	\end{minipage}
	\caption{Examples of our annotated data representing social groups of people in the scene and their common activity within their group. Realistically, individuals in the scene may perform their own actions, but they may also belong to a social group with a mutual activity, \textit{e.g.}, \textit{walking together}. In this figure, social groups and their corresponding common activities have been color-coded.}
	\label{fig:teaser}%\vspace{-1.5em}
\end{figure}
\\
\indent
In a real scenario, a scene may contain several people, individuals may perform their own actions while they might be connected to a social group. In other words, a real scene generally comprises several groups of people with potentially different social connections, \textit{e.g.} contribution toward a common activity or goal (Fig.~\ref{fig:teaser}). To this end, in our work we focus on the problem of ``\emph{Who is with whom and what they are doing together?}".
Although the existing works mentioned in the previous paragraphs tackle some elements, we propose a holistic approach that considers the multi-task nature of the problem, where these tasks are not completely independent, and which can benefit each other. Understanding of this scene-wide social context would be conducive for many video understanding applications, \textit{e.g.} anomalous behaviour detection in the crowd from a surveillance footage or navigation of an autonomous robot or car through a crowd. 
\\
\indent
To tackle this real-world problem, we propose an end-to-end trainable framework which takes a video sequence as input and learns to predict \textit{a}) each individual's action; \textit{b}) their social connections and groups; and, \textit{c}) a social activity for each predicted social group in the scene. We first introduce our framework for a relevant conventional problem, \textit{i.e.} group activity recognition~\cite{choi2011learning,deng2015deep,lan2011discriminative}. We propose an architecture design that incorporates: \emph{i}) \emph{I3D backbone}~\cite{gu2018ava} as a state-of-the art feature extractor to encode spatio-temporal representation of individuals in a video clip, \emph{ii}) \emph{Self-attention module}~\cite{vaswani2017attention} to refine individuals' feature representations, and \emph{iii}) \emph{Graph attention module}~\cite{velikovi2017graph} to directly model the interactions and connections among the individuals. Our framework outperforms the state-of-the-art on two widely adopted group activity recognition datasets. We then introduce our extended framework that can elegantly handle social grouping and social activity recognition for each group. We also augment an exisiting group activity dataset with enriched social group and social activity annotations for the social task. Our main contributions are:
    \begin{enumerate}
    \setlength{\itemsep}{1pt}
	\setlength{\parskip}{0pt}
	\setlength{\parsep}{0pt}
        \item We propose an end-to-end framework for the group activity recognition task through integration of I3D backbone, self-attention module and graph attention module in a well-justified architecture design. Our pipeline also outperforms existing solutions and sets a new state-of-the-art for the group activity recognition task on two widely adopted benchmarks.
        \item We show that by including a graph edge loss in the proposed group activity recognition pipeline, we obtain an end-to-end trainable solution to the multi-task problem of simultaneously grouping people, recognizing individuals' action and social activity of each social group~(social task). 
        \item We introduce new annotations, \textit{i.e.} social groups and social activity label for each sub-group on a widely used group activity dataset to serve as a new benchmark for the social task.
    \end{enumerate}

%% file: relatedwork.tex
\section{Related Work}
%\vspace{-.5em}
\textbf{Action Recognition.}
Video understanding is one of the main computer vision problems widely studied over the past decades. Deep convolutional neural networks (CNNs) have shown promising performance in action recognition on short trimmed video clips.
A popular approach in this line involves adoption of two-stream networks with 2D kernels to exploit the spatial and temporal information~\cite{feichtenhofer2016spatiotemporal,feichtenhofer2016convolutional,simonyan2014two,wang2016temporal}. Recurrent neural networks~(RNNs) have also been utilized to capture the temporal dependencies of visual features~\cite{donahue2015long,li2018videolstm}. Unlike these approaches,~\cite{ji20123d,tran2015learning} focused on CNNs with 3D kernels to extract features from a sequence of dense RGB frames. Recently,~\cite{carreira2017quo} proposed I3D, a convolutional architecture that is based on inflating 2D kernels pretrained on ImageNet into 3D ones. By pretraining on large-scale video datasets such as Kinetics~\cite{kay2017kinetics}, I3D outperforms the two-stream 2D CNNs on video recognition benchmark datasets.

\noindent
\textbf{Spatio-temporal Action Detection.} Temporal action detection methods aim to recognize humans' actions and their corresponding start and end times in untrimmed videos~\cite{sigurdsson2017asynchronous,sigurdsson2016hollywood,xu2017r,zhou2018temporal}. By introducing spatio-temporal action annotation for each subject, \textit{e.g.} as in AVA~\cite{gu2018ava}, spatio-temporal action detection received considerable attention~\cite{feichtenhofer2019slowfast,girdhar2018better,girdhar2019video,li2018recurrent,sun2018actor,wu2019long}. In particular,~\cite{sun2018actor} models the spatio-temporal relations to capture the interactions between human, objects, and context that are crucial to infer human actions. More recently, action transformer network~\cite{girdhar2019video} has been proposed to localize humans and recognize their actions by considering the relation between actors and context.

\noindent
\textbf{Group Activity Recognition.}
The aforementioned methods mostly focus on predicting individuals' actions in the scene. However, there are works focusing on group activity recognition where the aim is to predict a single group activity label for the whole scene. The early approaches typically extracted hand-crafted features and applied probabilistic graphical models~\cite{amer2014hirf,choi2012unified,choi2013understanding,choi2011learning,lan2012social,lan2011discriminative,shu2015joint} or AND-OR grammar models~\cite{amer2012cost,shu2015joint} for group activity recognition. In recent years, deep learning approaches especially RNNs achieve impressive performance largely due to their ability of both learning informative representations and modelling temporal dependencies in the sequential data~\cite{bagautdinov2017social,deng2016structure,deng2015deep,ibrahim2018hierarchical,ibrahim2016hierarchical,li2017sbgar,qi2018stagnet,ramanathan2016detecting,wang2017recurrent,shu2019hierarchical}. 
For instance,~\cite{ibrahim2016hierarchical} uses a two-stage LSTM model to learn a temporal representation of person-level actions and pools individuals' features to generate a scene-level representation. In~\cite{ramanathan2016detecting}, attention mechanism is utilized in RNNs to identify the key individuals responsible for the group activity. Later, authors of~\cite{shu2017cern} extended these works by utilizing an energy layer for obtaining more reliable predictions in presence of uncertain visual inputs.
Following these pipelines,~\cite{ibrahim2018hierarchical} introduces a relational layer that can learn compact relational representations for each person. The method proposed in~\cite{bagautdinov2017social} is able to jointly localize multiple people and classify the actions of each individual as well as their collective activity. In order to consider the spatial relation between the individuals in the scene, an attentive semantic RNN has been proposed in~\cite{qi2018stagnet} for understanding group activities. Recently, the Graph Convolutional Network (GCN) has been used in~\cite{wu2019learning} to learn the interactions in an Actor Relation Graph to simultaneously capture the appearance and position relation between actors for group activity recognition. Similarly~\cite{azar2019convolutional} proposed a CNN model encoding spatial relations as an intermediate activity-based representation to be used for recognizing group activities. There are also a number of attempts to simultaneously track multiple people and estimate their collective activities in multi-stage frameworks~\cite{choi2012unified,li2018did}. Although these approaches try to recognize the interactions between pairs of people by utilizing hand-crafted features, they are not capable of inferring social groups.

Despite the progress made towards action recognition, detection and group activity recognition~\cite{stergiou2019analyzing}, what still remains a challenge is simultaneously understanding of social groups and their corresponding social activity. Some existing works aim at only detecting groups in the scene \cite{choi2014discovering,ge2012vision} by relying on hand-crafted rules \textit{e.g.} face orientations, which are only applicable to very specific tasks such as conversational group detection~\cite{hung2011detecting,patron2010high,setti2013multi,swofford2019dante} and by utilizing small-scale datasets. In contrast to previous solutions which are task dependent and require domain expert knowledge to carefully design hand-crafted rules, we extend the concept of grouping to more general types of interactions. To this end, we propose an end-to-end trainable framework for video data to jointly predict individuals' action as well as their social groups and social activity of each predicted group.

%% file: method.tex
\section{Social Activity Recognition}
\begin{figure*}[!tbp]
  \centering
		\includegraphics[clip,width=1.0\linewidth]{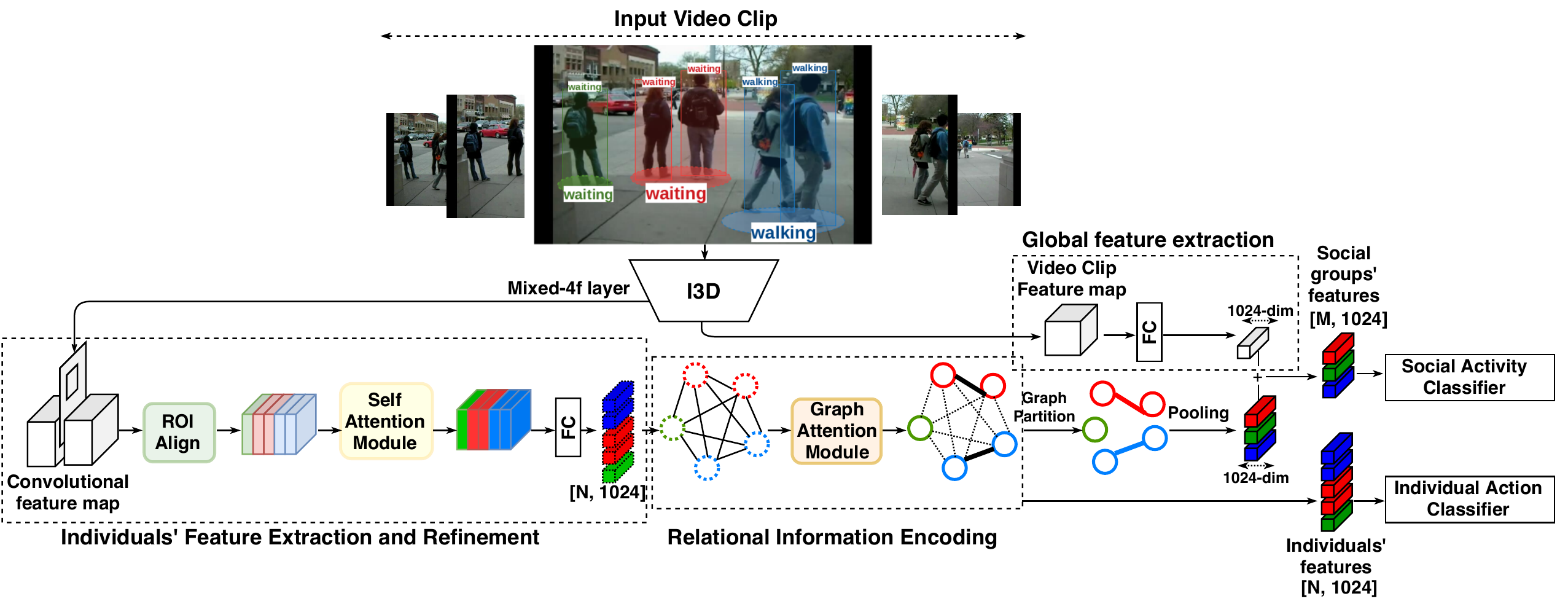}
\label{social_architecture}
\caption{Our network architecture for the social task. The set of aligned individuals' features are initially refined by the self-attention module. Projected feature maps are then fed into the GAT module to encode relational information between individuals. During training, feature representation of each social group is pooled from the feature maps of its members according to the ground-truth social connections. At test time, we adopt a graph partitioning algorithm on the inter-node attention coefficients provided by GAT and accordingly infer the social groups and social activity of each sub-group. M and N refer to the number of social groups and number of individuals respectively.}  
\label{fig: social_architecture}%\vspace{-1.0em}
\end{figure*}

Social activity recognition seeks to answer the question of ``\emph{Who is with whom and what they are doing together?}". Traditional group activity recognition can be seen as a simplified case of social activity recognition where all the individuals are assumed to form a single group in the scene and a single group activity label is predicted for the whole scene. For the ease of conveying our ideas, we present our framework first in the simpler setting of group activity recognition task, and then show how to augment it for the social task. 

\subsection{Group Activity Recognition Framework }\label{sec:gpframework}
%\vspace{-.5em}
A group activity recognition framework should be capable of: \textit{a}) generating a holistic and enriched spatio-temporal representation from the entire video clip; \textit{b}) extracting fine-detailed spatial features from bounding box of each person to accurately predict the individual actions; and \textit{c}) learning an aggregated representation from all individuals for precise realization of their collective activities. Illustrated in Fig.~\ref{fig: social_architecture}, we carefully design effective components in our framework to achieve the above desirable properties. We elaborate the components as follows.

\noindent
\textbf{I3D Backbone.} We use the Inflated 3D ConvNet (I3D)~\cite{carreira2017quo} (based on Inception architecture~\cite{ioffe2015batch}) as the backbone to capture the spatio-temporal context of an input video clip. In I3D, ImageNet pre-trained convolutional kernels are expanded into 3D, allowing it to seamlessly learn effective spatio-temporal representations. Motivated by the promising performance of the pre-trained I3D models in a wide range of action classification benchmarks, we exploit the feature representations offered by this backbone at multiple resolutions. More specifically, we use the deep spatio-temporal feature maps extracted from the final convolutional layer as a rich semantic representation describing the entire video clip. These deeper features provide low-resolution yet high-level representations that encode a summary of the video. 
In addition, accurate recognition of individuals' action rely on finer details which are often absent in very deep coarse representations. To extract fine spatio-temporal representations for the individuals, we use the higher resolution feature maps from the intermediate Mixed-4f layer of I3D. As depicted in Fig.~\ref{fig: social_architecture}, from this representation we extract the temporally-centered feature map corresponding to the centre frame of the input video clip. Given the bounding boxes in the centre frame, we conduct ROIAlign \cite{he2017mask} to project the coordinates on the frame's feature map and slice out the corresponding features for each individual's bounding box.

\noindent
\textbf{Self-attention Module.} 
Despite being localized to the individual bounding boxes, these representations still lack emphasis on visual clues that play a crucial role in understanding the underlying actions \textit{e.g.} a person's key-points and body posture. To overcome this, we adopt self-attention mechanism \cite{vaswani2017attention,nonlocal} to directly learn the interactions between any two feature positions of an individual's feature representation and accordingly leverage this information to refine each individual's feature map. In our framework the self-attention module functions as a non-local operation and computes the response at each position by attending to all positions in an individual's feature map. 
The output of the self-attention module contextualizes the input bounding box feature map with visual clues and thus, enriches the individual's representation by highlighting the most informative features. As substantiated by ablation studies in Section \ref{sec:group}, capturing such fine details significantly contribute to the recognition performance. 

\noindent
\textbf{Graph Attention Module.} Uncovering subtle interactions among individuals present in a multi-person scenario is fundamental to the problem of group activity recognition; each person individually performs an action and the set of inter-connected actions together result in the underlying global activity context. As such, this problem can elegantly be modeled by a graph, where the nodes represent refined individuals' feature map and the edges represent the interactions among individuals. We adopt the recently proposed Graph Attention Networks (GATs) \cite{velikovi2017graph} to directly learn the underlying interactions and seamlessly capture the global activity context. GATs flexibly allow learning attention weights between nodes through parameterized operations based on a self-attention strategy and have successfully demonstrated state-of-the-art results by outperforming existing counterparts~\cite{kipf2016semi}. GATs compute attention coefficients for every possible pair of nodes, which can be represented in an adjacency matrix $\hat{{O}}^{\alpha}$. 

\noindent
\textbf{Training.} In our framework, the GAT module consumes the individuals' feature map obtained from the self-attention component, encodes inter-node relations, and generates an updated representation for each individual. We acquire the group representation by max-pooling the enriched individuals' feature map and adding back a linear projection of the holistic video's features obtained from the I3D backbone. A classifier is then applied on this representation to generate group activity scores denoted by $\hat{{O}}^{\textrm{G}}$. Similarly, another classifier is applied on the individuals' representation to govern the individual action scores denoted by $\hat{{O}}^{\textrm{I}}_n$. The associated operations provide a fully differentiable mapping from the input video clip to the output predictions, allowing the framework to be trained in an end-to-end fashion by minimizing the following objective function,
\begin{equation}\label{eq:gp_loss}
    \mathcal{L} = \mathcal{L}_{\textrm{gp}}({O}^{\textrm{G}}, \hat{{O}}^{\textrm{G}}) + \lambda \sum _n\mathcal{L}_{\textrm{ind}}({O}^{\textrm{I}}_n, \hat{{O}}^{\textrm{I}}_n),
\end{equation}
where $\mathcal{L}_{\textrm{gp}}$ and $\mathcal{L}_{\textrm{ind}}$ respectively denote the cross-entropy loss for group activity and individual action classification. Here, ${O}^{\textrm{G}}$ and ${O}^{\textrm{I}}_n$ represent the ground-truth group activities and individual actions, $n$ identifies the individual and $\lambda$ is the balancing coefficient for the loss functions.   

\subsection{Social Activity Recognition Framework}
%\vspace{-.5em}
In a real-world multi-person scene, a set of social groups each with different number of members and social activity labels often exist. We refer to this challenging problem as the \textit{social activity recognition} task. In this section, we propose a novel yet simple modification to our group activity recognition framework that naturally allows understanding of social groups and their corresponding social activities in a multi-person scenario. The \textit{I3D backbone} and the \textit{self-attention module} remain exactly the same as elucidated in Section \ref{sec:gpframework}. We explain the required modifications for the \textit{graph attention module} as follows.

\noindent
\textbf{Training.} Previously, the GAT's inter-node attention coefficients were updated with the supervision signal provided by the classification loss terms in Eq. \ref{eq:gp_loss}. To satisfy the requirements of the new problem, \textit{i.e.} to generate social groups and their corresponding social activity label, we augment the training objective with a graph edge loss $\mathcal{L}_c$ that incentivizes the GAT's self-attention strategy to converge to the individuals' social connections
\setlength{\belowdisplayskip}{10pt} \setlength{\belowdisplayshortskip}{10pt}
\setlength{\abovedisplayskip}{10pt} \setlength{\abovedisplayshortskip}{10pt}
\begin{equation}
\begin{aligned}
    \mathcal{L} = &  \sum_s\mathcal{L}_{\textrm{sgp}}({O}^{\textrm{SG}}_s, \hat{{O}}^{\textrm{SG}}_s) +
    \lambda_1 \sum _n\mathcal{L}_{\textrm{ind}}({O}^{\textrm{I}}_n, \hat{{O}}^{\textrm{I}}_n) +  
    \lambda_2\mathcal{L}_{\textrm{c}}({O}^\alpha,\hat{{O}}^\alpha),
\label{eq:sgp_loss}
\end{aligned}
\end{equation}
where, $\mathcal{L}_{c}$ is the binary cross-entropy loss to reduce the discrepancy between GAT's learned adjacency matrix $\hat{{O}}^{\alpha}$ and the ground-truth social group connections ${O}^{\alpha}$. Further, $\mathcal{L}_{\textrm{sgp}}$ and $\mathcal{L}_{\textrm{ind}}$ respectively denote the cross-entropy loss for social activity of each social group and individual actions classification. Notably, given the ground-truth social groupings during training, we achieve the representation for each social group by max-pooling its corresponding nodes' feature-map and adding back a linear projection of the video features obtained from the I3D backbone (similar to learning group activity representations in our simplified group activity framework). A classifier is then applied on top to generate the social activity scores $\hat{{O}}^{\textrm{SG}}_s$. At inference time however, we require a method to infer the social groups in order to compute the corresponding social representations. To this end, we propose to utilize graph spectral clustering \cite{zelnik2005self} on the learned attention coefficients by GAT, $\hat{{O}}^{\alpha}$ and achieve a set of disjoint partitions representing the social groups. In the above formulation $s$ is the social group identifier and ($\lambda_1$, $\lambda_2)$ are the loss balancing coefficients.

%% file: dataset.tex
\section{Datasets}
%\vspace{-.5em}
We evaluate our group activity recognition framework on two widely adopted benchmarks: Volleyball dataset and Collective Activity dataset~(CAD). We also perform evaluation of our social activity recognition framework on our provided social task dataset by augmenting the CAD with social groups and social activity label for each group annotations.

\subsection{Group Activity Recognition Datasets}
\textbf{Volleyball Dataset~\cite{ibrahim2016hierarchical}} contains 4830 videos from 55 volleyball games partitioned into 3493 clips for training and 1337 clips for testing. Each video has a group activity label from the following activities: \textit{right set, right spike, right pass, right win-point, left set, left spike, left pass, left win-point}. The centered frame of each video is annotated with players' bounding boxes and their individual action including \textit{waiting, setting, digging, failing, spiking, blocking, jumping, moving, standing}. 

\noindent
\textbf{Collective Activity Dataset(CAD)~\cite{choi2009they}} has 44 video sequences captured from indoor and street scenes. In each video, actors' bounding box, their actions and a single group activity label are annotated for the key frames (\textit{i.e.} 1 frame out of every 10 frames). Individual actions include \textit{crossing, waiting, queuing, walking, talking, N/A}. The group activity label associated with each key frame is assigned according to the majority of individuals' actions in the scene. We adopt the same train/test splits as previous works \cite{azar2019convolutional,qi2018stagnet,wu2019learning}.

\subsection{Social Activity Recognition Dataset}
\label{subsec:social}
%\vspace{-.5em}
%______________________________________________________________________
%\begin{figure}[t]
%  \centering
%	\includegraphics[width=.6\linewidth]{chart.pdf}
%	\caption{The histogram of social activities for varying social group sizes (1 to 4 people per social group).}
%	\label{fig:dataset_statistics}
%\end{figure}

\begin{wrapfigure}{L}{0.5\textwidth}
  \centering
	\includegraphics[width=0.5\textwidth]{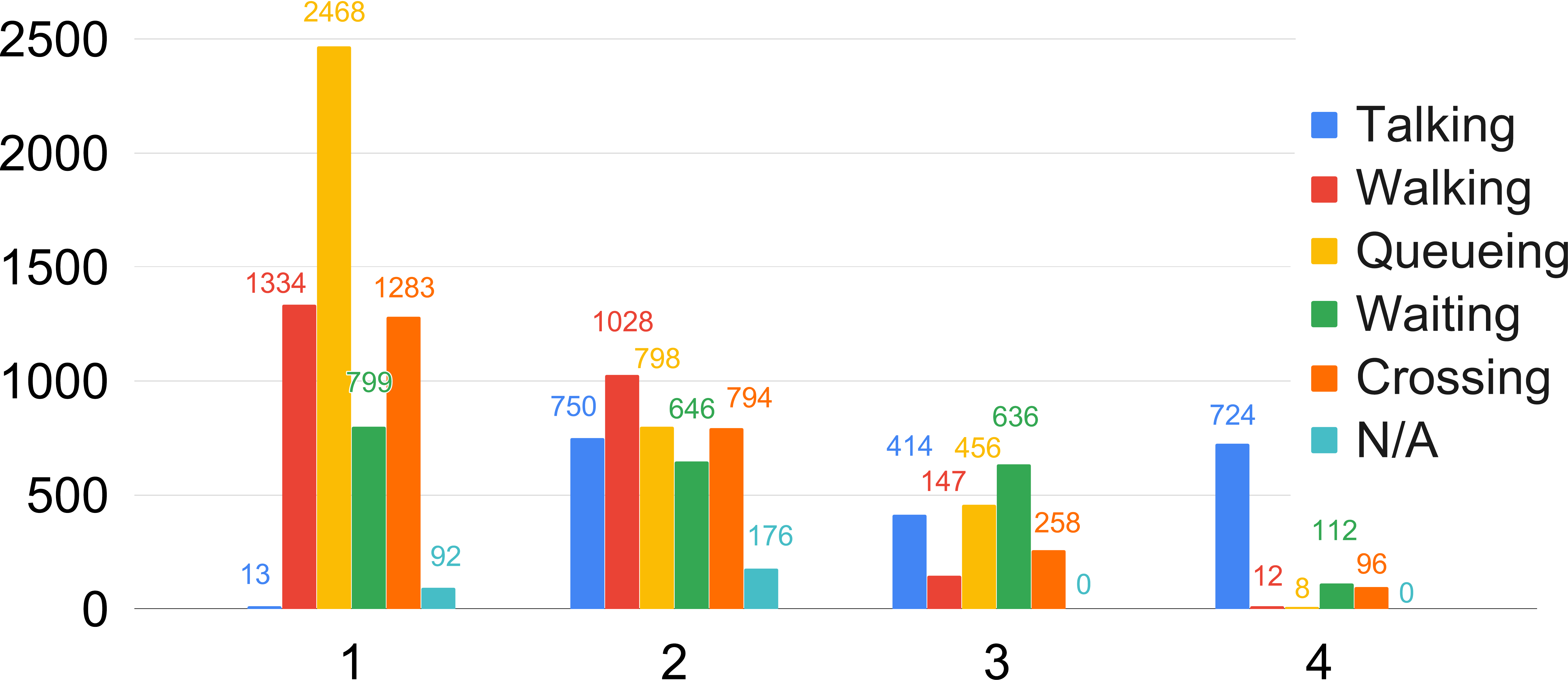}
	\caption{The histogram of social activities for varying social group sizes (1 to 4 people per social group).}
	\label{fig:dataset_statistics}
\end{wrapfigure}

In order to solve the problem of social activity recognition, a video dataset containing scenes with different social groups of people, each performing a social activity is required. Thus, we decided to utilize CAD which is widely used in the group activity recognition task and its properties suit well our problem. Other video action datasets~\cite{kay2017kinetics,gu2018ava,joo2017panoptic} could not be used in this problem since they mostly consist of scenes with only one social group or a number of sigleton groups. We provide enriched annotations on CAD for solving the social task, which we call \emph{Social-CAD}. As such, for each key frame, we maintain the exact same bounding box coordinates and individual action annotations as the original dataset. However, rather than having a single group activity label for the scene, we annotate different social groups and their corresponding social activity labels. Since there may only exist a subtle indicator in the entire video sequence, \textit{e.g.} an eye contact or a hand shake, suggesting a social connection between the individuals, determining \emph{"who is with whom"} can be a challenging and subjective task. Therefore, to generate the social group annotations as reliable as possible: \textit{a}) we first annotate the trajectory of each person by linking his/her bounding boxes over the entire video, \textit{b}) Given the trajectories, we asked three annotators to independently divide the tracks into different sub-groups according to their social interactions, and \textit{c}) we adopted majority voting to confirm the final social groups. Similar to the CAD annotation, the social activity label for each social group is defined by the dominant action of its members. We use the same train/test splits in Social-CAD as in CAD. Detailed activity distributions of Social-CAD are given in Fig~\ref{fig:dataset_statistics}.

%% file: experiment.tex
\section{Experimental Results}
%\vspace{-.5em}
To evaluate our proposed framework, we first show that our group activity recognition pipeline outperforms the state-of-arts in both individual action and group activity recognition tasks on two widely adopted benchmarks. Then we evaluate the performance of our social activity recognition framework on Social-CAD.
  
\subsection{Group Activity Recognition}\label{sec:group}
%\vspace{-.5em}
\begin{figure}[b]
  \centering
	\includegraphics[clip,width=1.0\linewidth]{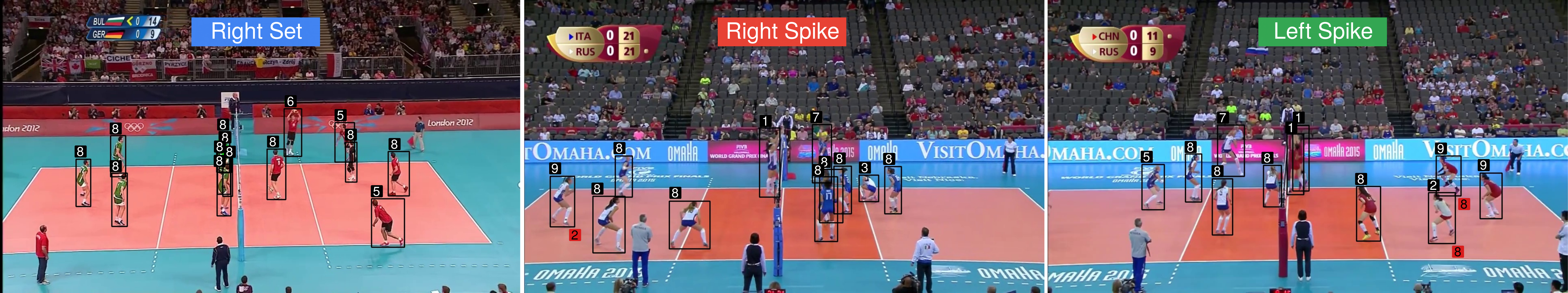}%\vspace{-1em}
	\caption{Visual results of our method on Volleyball dataset for the individual/group activity recognition task~(better viewed in color). The bounding boxes around players are produced by our detection-based approach. The numbers above the boxes denote the predicted action IDs. The ground-truth action ID for each player is indicated in red when the predicted action ID is wrong. The label on top of each key frame shows the predicted group activity. Please refer to the dataset section to map IDs to their corresponding actions.}
	\label{fig:volleyball_vis}%\vspace{-1.0em}
\end{figure}

\begin{figure}%[bt]
  \centering
	\includegraphics[width=\textwidth]{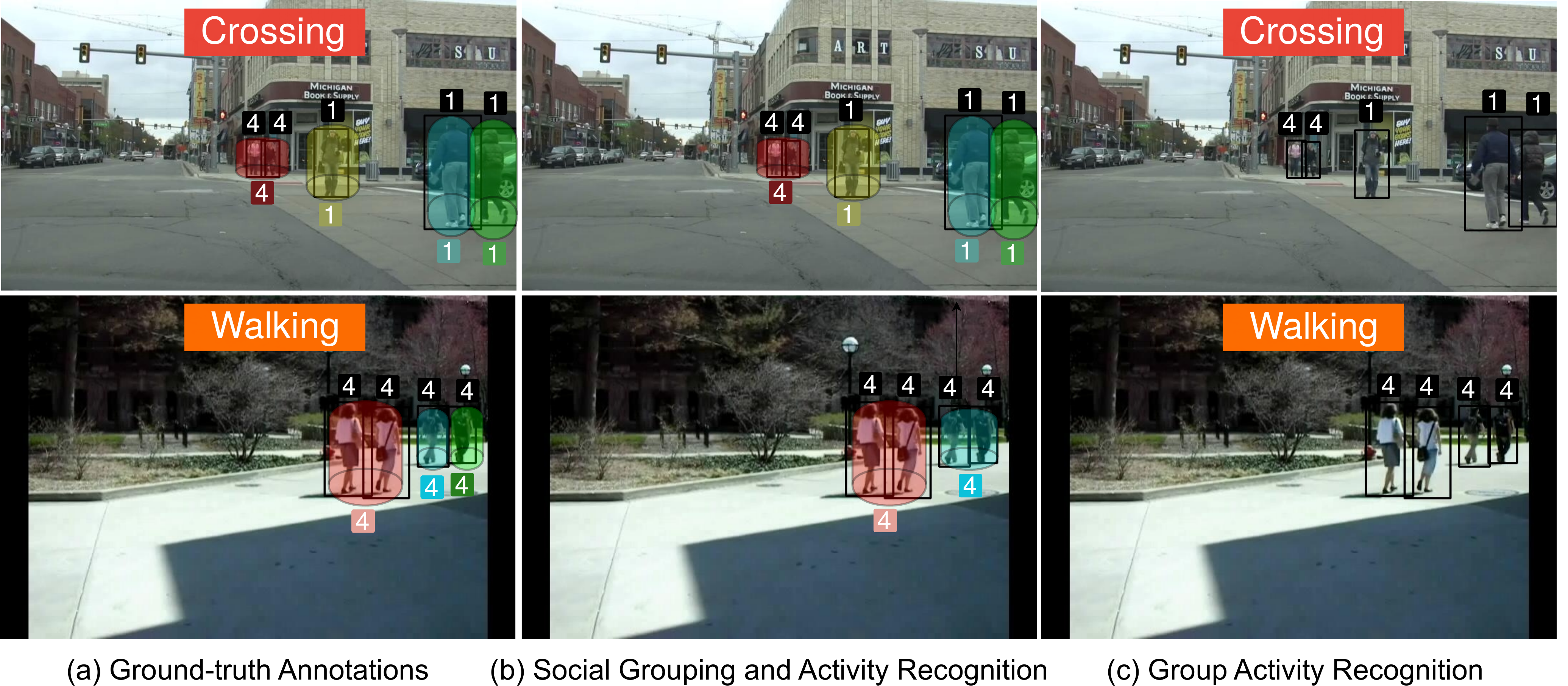}%\vspace{-1em}
	\caption{Visual results of our method on CAD for both social and group activity recognition tasks. Column(a) shows the ground-truth annotation for both tasks. Column(b) represents our prediction for the social task. Column(c) is our predictions for the group task. Note that social groups are denoted by a colored cylinder with their social group labels underneath. The numbers on top of bounding boxes denote the individual action IDs and the label tag above each key frame is the group activity label for the whole scene. 1 and 4 refer to crossing and walking activities respectively.}
	\label{fig:social_CAD_vis}%\vspace{-5 pt}
\end{figure}

\noindent
\textbf{Implementation Details.}
In our model, we use an I3D backbone which is initialized with Kintetics-400~\cite{kay2017kinetics} pre-trained model. We utilize ROI-Align with crop size of $5 \times 5$ on extracted feature-map from Mixed-4f layer of I3D. We perform self-attention on each individuals' feature map with query, key and value being different linear projections of individuals' feature map with output sizes being $1/8, 1/8, 1$ of the input size. We then learn a 1024-dim feature map for each individual features obtained from self-attention module. Aligned individuals' feature maps are fed into our single-layer, multi-head GAT module with 8 heads and input-dim, hidden-dim, output-dim being 1024 and droupout probability of 0.5 and $\alpha=0.2$ \cite{velikovi2017graph}. 
We utilize ADAM optimizer with $\beta_1 = 0.9$, $\beta_2 = 0.999$, $\epsilon = 10^{-8}$ following \cite{wu2019learning}. We train the network in two stages. First, we train the network without the graph attention module. Then we fine-tune the network end-to-end including GAT.
For the Volleyball dataset, we train the network in 200 epochs with a mini-batch size of 3 and a learning rate ranging from $10^{-4}$ to $10^{-6}$ and $\lambda_1=8$. For the CAD, we train the network in 150 epochs with a mini-batch size of 4 and a learning rate ranging from $10^{-5}$ to $10^{-6}$ and $\lambda_1=10$. Input video clips to the model are 17 frames long, with the annotated key frame in the centre. At test time, we perform our experiments based on two widely used settings in group activity recognition literature namely groundtruth-based and Detection-based settings.%In both settings, ground-truth bounding boxes are utilized in training. 
In the groundtruth-based setting, ground-truth bounding boxes of individuals are given to the model to infer the individual action for each box and the group activity for the whole scene. In the detection-based setting, we fine-tune a Faster-RCNN \cite{ren2015faster} on both datasets and utilize the predicted boxes for inferring the individuals' action and group activity.

\noindent
\textbf{Evaluation.} In order to evaluate the performance of our model for the group activity recognition task, we adopt the commonly used metric, \textit{i.e.} average accuracy, reported in all previous works~\cite{azar2019convolutional,ibrahim2018hierarchical,qi2018stagnet,wu2019learning}. To report the performance of individuals' action in GT-based setting, similar to~\cite{wu2019learning}, we used the average accuracy as the measure. In the case of Detection-based setting, average accuracy for evaluating the individuals action is not a valid measure due to the presence of false and missing detections. Instead, we report the commonly used measure for object detection, \textit{i.e.} mean average precision (mAP)~\cite{everingham2010pascal}.     

\noindent
\textbf{Ablation Study.}
We justify the choice of each component in our framework with detailed ablation studies. The results on volleyball and CAD are shown in Table \ref{table:volleyball+collective_ablation}. As the simplest baseline denoted by Ours[group-only], we use a Kinetics-400 pre-trained I3D backbone and fine-tune it by utilizing the input videos' feature representation obtained from the last layer of I3D and using a cross-entropy loss to learn the group activity without considering individuals' action. It is worth mentioning that surprisingly, our simplest baseline already outperforms many existing frameworks on group activity recognition (see the group accuracy in Table~\ref{table:volleyball+collective_group}). This shows the importance of extracting spatio-temporal features simultaneously using 3D models as well as taking into account the whole video clip rather than solely focusing on individuals and their relations. 
To consider the effect of jointly training the model on group activity and individual action tasks, we add a new cross-entropy loss to our simplest baseline for training the individuals' action. As Ours[w/o SA- w/o GAT] experiment shows, training both tasks jointly helps improve the group activity recognition performance. In Ours[w/o GAT], we add the self-attention module performing on each individual's feature-map in order to highlight the most important features and improve the individual action recognition performance. As shown in Ours[w/o GAT], utilizing self-attention module improves the individual action accuracy by 1.2\% on volleyball dataset and by 2.1\% on CAD which also contributes to a slight improvement in the group activity accuracy on both datasets. Finally, we add the GAT module to capture the interactions among individuals which is essential in recognizing group activity. As shown in the Ours[final] experiment, utilizing GAT increases the group activity accuracy by 0.6\% on volleyball and by 1.1\% on collective activity dataset. GAT also improves the individual action performance on both volleyball and collective active datasets by 0.1\% and 3\% respectively. The higher boost in individual action performance on CAD compared to the one in the volleyball dataset shows the effectiveness of GAT in highlighting social sub-groups and updating individual feature representations accordingly as it is benefiting from a self attention strategy between nodes. 
%_________________________________ABLATION______________________________________

\begin{table}[t]
\begin{center}
\caption{Ablation study of our method for group activity recognition. w/o SA: without self-attention module. w/o GAT: without graph attention module.}%\vspace{-.5em}
\footnotesize
\begin{tabular}{l|c||c|}
\cline{2-3}
& \multicolumn{1}{c||}{Volleyball}&\multicolumn{1}{c|}{Collective Activity} \\                        
 \cline{2-3} 
& \begin{tabular}[c]{@{}c@{}}Group (Individual)\\ \textbf{Acc.\% (Acc.\%)}\end{tabular} & \begin{tabular}[c]{@{}c@{}}Group (Individual)\\ \textbf{Acc.\% (Acc.\%)}\end{tabular}                                                 \\ \hline

\multicolumn{1}{|l|}{Ours{[}group-only{]}}      & 91.0 (-)                                                               &  84.6 (-)                                                                                                                                            \\ \hline
\multicolumn{1}{|l|}{Ours{[}w/o SA- w/o GAT{]}} & 92.0 (82.0) & 88.2 (73.4)                                                                                                                               \\ \hline
\multicolumn{1}{|l|}{Ours{[}w/o GAT{]}}         & 92.5 (83.2) & 88.3 (75.3)                                                                                                                             \\ \hline
\multicolumn{1}{|l|}{\textbf{Ours{[}final{]}}}           & \textbf{93.1(83.3)} & \textbf{89.4 (78.3)}                                                                                 \\ \hline
\end{tabular}%\vspace{-.8em}
\label{table:volleyball+collective_ablation}
\end{center}
\end{table}
%______________________________GROUP_TASK___________________________________
\begin{table}%[t]
\footnotesize
\centering
\caption{Comparison with the state-of-the-arts on Volleyball dataset and CAD for group activity recognition.}%\vspace{-.5em}
\begin{tabular}{l|c||c|}
\cline{2-3}
        & Volleyball                                                     & Collective Activity   \\ \cline{2-3} 
& \begin{tabular}[c]{@{}c@{}}Group (Individual)\\ Acc.\% (Acc.\%)\end{tabular} & \begin{tabular}[c]{@{}c@{}}Group (Individual)\\ Acc.\% (Acc.\%)\end{tabular}    \\ \hline
\multicolumn{1}{|l|}{HDTM~\cite{ibrahim2016hierarchical}}         & 81.9 (-)                                                                                                                                  & 81.5 (-)                                                                  \\ \hline
\multicolumn{1}{|l|}{CERN~\cite{shu2017cern}}         & 83.3 (-)                                                                                                                                   & 87.2 (-)                                                                   \\ \hline
\multicolumn{1}{|l|}{StagNet~\cite{qi2018stagnet}}  & 89.3 (-)                                                                                                                                & 89.1 (-)                                                                   \\ \hline
\multicolumn{1}{|l|}{HRN~\cite{ibrahim2018hierarchical}}          & 89.5 (-)                                                                                                                                 & - (-)                                                                      \\ \hline
\multicolumn{1}{|l|}{SSU~\cite{bagautdinov2017social}}      & 90.6 (81.8)                                                                                                                            & - (-)                                                                      \\ \hline
\multicolumn{1}{|l|}{CRM~\cite{azar2019convolutional}}          & 93.0 (-)                                                                                                                               & 85.8 (-)                                                                                  \\ \hline
\multicolumn{1}{|l|}{ARG$\dagger$~\cite{wu2019learning}}     & 92.5 (83.0)                                                                                                                         & 88.1 (77.3)                                                                   \\ \hline
\multicolumn{1}{|l|}{\textbf{Ours}}     & \textbf{93.1} (\textbf{83.3})                                                                                      & \textbf{89.4} (\textbf{78.3})                           
                                \\
\hline\hline
& Acc.\% (mAP\%) & Acc.\% (mAP\%) \\ \hline
\multicolumn{1}{|l|}{StagNet(Det)~\cite{qi2018stagnet}} & 87.6 (-)                                                                                                                                  & 87.9 (-)                                                                   \\ \hline                                
\multicolumn{1}{|l|}{SSU(Det)~\cite{bagautdinov2017social}}     & 86.2 (-)                                                                                                                          & - (-)    \\ \hline                           
\multicolumn{1}{|l|}{ARG(Det)$\dagger$~\cite{wu2019learning}}     & 91.5 (39.8)                                                                                                                                   & 86.1 (49.6)                                                                       \\ \hline
\multicolumn{1}{|l|}{\textbf{Ours(Det)}}    & \textbf{93.0 (41.8)}                                                                                                          & \textbf{89.4 (55.9)}  \\\hline
\end{tabular}%\vspace{-5 pt}
\label{table:volleyball+collective_group}
\end{table}
%______________________________GROUP_TASK___________________________________
\noindent
\textbf{Comparison with the State-of-the-arts.}
We compare our results on Volleyball and CAD with the state-of-the-art methods in Table \ref{table:volleyball+collective_group}, using group activity accuracy and individual action accuracy as the evaluation metrics. The top section of the table demonstrates the performance of the approaches in the groundtruth-based setting, where ground-truth bounding box of each person is used for prediction of the individual action as well as group activity of the whole scene. However, in the detection-based settings (indicated by (Det) in Table~\ref{table:volleyball+collective_group}), a Faster-RCNN is fine-tuned on both datasets and predicted bounding boxes for individuals are used during inference (which is more realistic in practice). In group activity recognition using predicted bounding boxes, our model has the least performance drop compared to other methods. In Table \ref{table:volleyball+collective_group}, ARG$\dagger$ is the result that we obtained by running \cite{wu2019learning}'s released code with the same setting mentioned for each dataset, and the reproduced results perfectly matched the reported results on volleyball dataset. However, we could not reproduce their reported results on CAD. Having their source code available, in order to have a fair comparison with our framework, we also reported their performance on individuals' action on CAD datasets using both groundtruth-based and detection-based settings (not reported in the original paper). Our framework outperforms all existing methods in all settings on both datasets. We observe that a common wrong prediction in all the existing methods on CAD is the confusion between \emph{crossing} and \emph{walking} in the previous setting. \emph{crossing} and \emph{walking} are essentially same activities being performed at different locations. Thus, we merge these two classes into a single \emph{moving} class and report the Mean Per Class Accuracy (MPCA) in Table \ref{table:MPCA} as in \cite{azar2019convolutional}. ARG$\dagger$ outperforms our method in one class \emph{moving}, and our model performs the best in all other 3 classes and the overall metric MPCA.     

\begin{table}[t]
\footnotesize
\centering
\caption{The mean per class group activity accuracies (MPCA) and per class group activity accuracies of our model compared to the existing models on CAD. M, W, Q and T stand for Moving, Waiting, Queuing and Talking respectively. Note that Crossing and Walking are merged as Moving.}
\begin{tabular}{|l|c|c|c|c|c|}
\hline
Method   & M     & W             & Q              & T              & MPCA          \\ \hline
HDTM~\cite{ibrahim2016hierarchical}     & 95.9  & 66.4          & 96.8           & 99.5           & 89.7          \\ \hline
SBGAR~\cite{li2017sbgar}    & 90.08 & 81.4          & 99.2           & 84.6           & 89.0          \\ \hline
CRM~\cite{azar2019convolutional}      & 91.7  & 86.3          & 100.0          & 98.91          & 94.2          \\ \hline
ARG$\dagger$~\cite{wu2019learning}      & \textbf{100.0}  & 76.0          & 100.0          & 100.0          & 94.0          \\ \hline
\textbf{Ours}     & 98.0  & \textbf{91.0} & \textbf{100.0} & \textbf{100.0} & \textbf{97.2} \\ \hline
\end{tabular}
%\vspace{-.8em}
\label{table:MPCA}
\end{table}
%_________________________________SOCIAL________________________________________________________
\subsection{Social Activity Recognition}
%\vspace{-.5em}
\textbf{Implementation Details.}
Our model is trained end-to-end for 200 epochs with a mini-batch size of 4 and a learning rate of $10^{-5}$ and $\lambda_1 = 5$ and $\lambda_2 = 2$. Other hyper-parameters have the same values as in the group activity recognition experiments on CAD. For graph partitioning at test time, we used graph spectral clustering technique~\cite{ng2002spectral,zelnik2005self}.

\noindent
\textbf{Evaluation.} Similar to the group activity recognition task, we perform experiments in two groundtruth-based and detection-based settings. For the social task, we evaluate three sub-tasks: 1) \textit{social grouping}, 2) \textit{social activity recognition of each group} and 3) \textit{individuals' action recognition}. In the GT-based setting, for (1) we calculate the \textit{membership accuracy} as the accuracy of predicting each person's assignment to a social group~(including singleton groups). This accuracy is also known as unsupervised clustering accuracy~\cite{unacc}. For (2), we evaluate if both the membership and the social activity label of a person are jointly correct. If so, we consider this instance as a true positive, otherwise, it is assumed a false positive. The final measure, \textit{i.e.} \textit{social accuracy} is attained as a ratio between the number of true positives and the number of predictions. For (3), we evaluate if the individual's action label is correctly predicted and report \textit{individual action accuracy}. In the detection-based setting, mAP is reported in order to evaluate predicted sub-groups, social activity of each group and individuals' action. For this experiment, bounding boxes with N/A groundtruth are excluded.

\noindent
\textbf{Results and Comparison.}
The performance of our method on social task is reported in Table \ref{table:social}, using membership accuracy, social activity accuracy and individual action accuracy in GT-based setting and mAP for each sub-task in detection-based setting as evaluation metrics. We consider three scenarios of baselines for evaluating this task: 
\\
\noindent
\textit{1)~Single group setting:} forcing all individuals form a single social group, and then assessing algorithms' performance. ARG[group]~\cite{wu2019learning} and ours[group] essentially are the approaches in Table~\ref{table:volleyball+collective_group}, but are evaluated in membership and social activity metrics. GT[group] uses ground-truth activity labels serving as the upper bound performance for group activity recognition frameworks;
\\
\noindent
\textit{2)~Individuals setting:} forcing each individual as a unique social group, \textit{e.g.}, if there are 10 people in the scene, they will be considered as 10 social groups. GT[individuals] uses ground-truth action labels serving as the upper bound performance for group activity recognition frameworks;
\\
\noindent
\textit{3)~Social group setting:} Partitioning individuals into multiple social groups. Our first approach uses group activity recognition pipeline in training and uses graph spectral clustering technique to find social groups at inference, named as Ours{[}cluster{]} (third part of Table~\ref{table:social}). This produces better performance compared to the other baselines, but it is outperformed by our final framework denoted by Ours{[}learn2cluster{]}, where we learn representations via the the additional graph edge loss.
%_______________________________________________________________________________________
\begin{table}[t]
%\begin{adjustbox}{width=80mm}
\centering
\caption{Social activity recognition results. In each column we report accuracy and mAP for the groundtruth-based and detection-based settings respectively.}%\vspace{-.5em}
\footnotesize
\begin{tabular}{l|c|c|c|}
\cline{2-4}
                                                                    & \begin{tabular}[c]{@{}c@{}}Membership\\ \hline GT(Det) \\ Acc.\%,(mAP\%)\end{tabular} & \begin{tabular}[c]{@{}c@{}}Social Activity\\ \hline GT(Det) \\ Acc.\%, (mAP\%)\end{tabular}&
                                                                    \begin{tabular}[c]{@{}c@{}}Individual Action\\ \hline GT(Det) \\ Acc.\%, (mAP\%)\end{tabular}
                                                                    
                                                                    \\ \hline
\multicolumn{1}{|l|}{ARG{[}group{]}~\cite{wu2019learning}}                                           & 54.4(49.0)                                                          & 47.2(34.8)           & 78.4(62.6)                                                    \\ \hline
\multicolumn{1}{|l|}{Ours {[}group{]}}                               & 54.4(49.0)                                                          & 47.7(35.6) & 79.5(64.2)                                                                \\ \hline
\multicolumn{1}{|l|}{GT{[}group{]}}                                & 54.4(-)                                                          & 51.6(-)  & -(-)                                                              \\ \hline\hline
\multicolumn{1}{|l|}{ARG{[}individuals{]}~\cite{wu2019learning}}                                           & 62.4(52.4)                                                          & 49.0(41.1)             & 78.4(62.6)                                                 \\ \hline
\multicolumn{1}{|l|}{Ours{[}individuals{]}}                               & 62.4(52.4)                                                          & 49.5(41.8) & 79.5(64.2) \\ \hline
\multicolumn{1}{|l|}{GT{[}individuals{]}}                                & 62.4(-)                                                          & 54.9(-)        & -(-)                                                       \\ \hline\hline
\multicolumn{1}{|l|}{Ours{[}cluster{]}}   & 78.2(68.2)                                                          & 52.2(46.4)     & 79.5(64.2)                                                               \\ \hline

\multicolumn{1}{|l|}{Ours{[}learn2cluster{]}}             & \textbf{83.0(74.9)}                                                          & \textbf{69.0(51.3)}    & \textbf{83.3(66.6)}                                                               \\ \hline
\end{tabular}
%\vspace{-.8em}
\label{table:social}
\end{table}

%___________________________________________________________________________________________
\noindent
\textbf{Discussion.} As mentioned in Section~\ref{subsec:social}, there might exist only a single frame with a subtle gesture in the entire video, demonstrating a social connection between people in the scene. Therefore as future direction, incorporating individuals' track and skeleton pose might disambiguate some challenging cases for social grouping. Moreover, the performance of our proposed social grouping framework heavily relies on the performance of the graph spectral clustering technique, which is not part of the learning pipeline. Improving this step by substituting it with a more reliable graph clustering approach, or making it a part of learning pipeline can potentially ameliorate the final results.

%% file: conclusion.tex
\section{Conclusion}
%\vspace{-.5em}
In this paper, we propose the novel social task which requires jointly predicting of individuals' action, grouping them into social groups based on their interactions and predicting the social activity of each social group. To tackle this problem, we first considered addressing the simpler task of group activity recognition, where all the individuals are assumed to form a single group and a single group activity label is predicted for the scene. As such, we proposed a novel deep framework incorporating well-justified choice of state-of-the-art modules such as I3D backbone, self-attention and graph attention network. We demonstrate that our proposed framework achieves state-of-the-art results on two widely adopted datasets for the group activity recognition task. Next, we introduced our social task dataset by providing additional annotations and re-purposing an existing group activity dataset. We discussed how our framework can readily be extended to handle social grouping and social activity recognition of groups through incorporation of a graph partitioning loss and a graph partitioning algorithm \textit{i.e.} graph spectral clustering. In the future, we aim to use the social activity context for development of better forecasting models, \textit{e.g.} the task of social trajectory prediction, or social navigation system for an autonomous mobile robot.